\documentclass[11pt,a4paper]{article}
\usepackage{lipsum}
\usepackage[hyperref]{acl2017}
\usepackage{times}
\usepackage{latexsym}
\usepackage{graphicx,subfigure}
\usepackage{amsmath}
\usepackage{amsfonts}
\usepackage{multirow}
\usepackage[linesnumbered,ruled]{algorithm2e}
\usepackage{enumitem}

\setlist[itemize]{noitemsep, topsep=0pt}

\usepackage{url}

\aclfinalcopy 

\title{Representing Sentences as Low-Rank Subspaces}

\author{Jiaqi Mu, \ Suma Bhat, \and Pramod Viswanath \\
University of Illinois at Urbana Champaign\\
{\tt \{jiaqimu2, spbhat2, pramodv\}@illinois.edu}}

\date{}

\begin{document}
\maketitle
\begin{abstract} 
Sentences are important semantic units of natural language. A generic, distributional  representation of sentences that can capture the latent semantics is beneficial to multiple downstream applications. We observe a simple  geometry of sentences -- the word representations of a given sentence (on average 10.23 words in all SemEval datasets with a standard deviation 4.84) roughly lie in a {\em low-rank} subspace (roughly, rank 4). Motivated by this observation, we represent a sentence by the  low-rank {\em subspace} spanned by its word vectors. Such an unsupervised representation is empirically validated via  semantic textual similarity tasks on 19 different datasets, where it outperforms the sophisticated neural network models,  including skip-thought vectors, by 15\% on average. 
\end{abstract}

\section{Introduction}

Real-valued word representations have brought a fresh approach to classical problems in NLP, recognized for their ability to capture linguistic regularities: similar words tend to have similar representations; similar word pairs tend to have similar difference vectors 
\cite{bengio2003neural,mnih2007three,mikolov2010recurrent,collobert2011natural,huang2012improving,dhillon2012two,mikolov2013efficient,pennington2014glove,levy2014neural,arora2015rand,stratos2015model}. Going beyond words, sentences capture much of the semantic information. Given the success of lexical representations, a natural question of great topical interest is how to extend the power of distributional representations to sentences. 

There are currently two approaches to represent sentences. A sentence contains rich syntactic information and can be modeled through sophisticated neural networks (e.g., convolutional neural networks \cite{kim2014convolutional,kalchbrenner2014convolutional}, recurrent neural networks \cite{sutskever2014sequence,le2014distributed,kiros2015skip,hill2016learning} and recursive neural networks \cite{socher2013recursive}). Another simple and common approach ignores the latent structure of sentences: a prototypical approach  is to represent a sentence by summing or averaging over the vectors of the words in this sentence \cite{wieting2015towards,adi2016fine,kenter2016siamese}. 

Recently, \citet{wieting2015towards,adi2016fine} reveal that even though the latter  approach ignores all syntactic information, it is simple, straightforward, and remarkably robust at capturing the sentential semantics. Such an approach successfully outperforms the neural network based approaches on textual similarity tasks in both {\rm supervised} and {\rm unsupervised} settings.

We follow the latter approach but depart from representing sentences in a vector space as in these prior works; we  present a novel Grassmannian property of sentences. The geometry is motivated  by \cite{gong2016geometry,mu2016geometry} where an interesting phenomenon is observed -- the local context of a given word/phrase can be well represented by a {\rm low rank subspace}. We propose to generalize this observation to sentences: not only do the word vectors in a snippet of a sentence (i.e., a context for a given word defined as several words surrounding it) lie in a low-rank subspace, but the entire sentence (on average 10.23 words in all SemEval datasets with standard deviation 4.84) follows this geometric property as well:
\begin{quote}
{\bf  Geometry of Sentences:} The word representations lie in a low-rank subspaces (rank 3-5) for all words in a target sentence. 
\end{quote}

The observation indicates that the subspace contains most of the information about this sentence, and therefore motivates a sentence representation method: the sentences should be represented in the space of subspaces (i.e., the Grassmannian manifold) instead of a  vector space;  formally: 
\begin{quote}
{\bf Sentence Representation:} A sentence can be represented by a low-rank subspace spanned by its  word representations.
\end{quote}

Analogous to word representations of similar words being similar vectors, the principle of sentence representations is: similar sentences should  have similar subspaces. Two questions arise: (a) how to define the similarity between sentences and (b) how to define the similarity between subspaces. 

The first question has been already addressed by the popular semantic textual similarity (STS) tasks. Unlike textual entailment (which aims at inferring directional relation between two sentences) and paraphrasing (which is a binary classification problem), STS provides a unified framework of measuring the degree of semantic equivalence \cite{agirre2012semeval,agirre2013sem,agirre2014semeval,agirrea2015semeval} in a continuous fashion. Motivated by the cosine similarity between vectors being a good index for word similarity, we generalize this metric to subspaces: the similarity between subspaces defined in this paper is the $\ell_2$-norm of the singular values between two subspaces; note that the singular values are in fact the cosine of the principal angles.  

The key justification for our approach comes from empirical results that outperform state-of-the-art in some cases, and being comparable in others. In summary, representing sentences by subspaces outperforms representing sentences by averaged word vectors (by 14\% on average) and sophisticated neural networks (by 15\%) on 19 different STS datasets, ranging over different domains (News, WordNet definition, and Twitter). 

\section{Geometry of Sentences}

Assembling successful distributional word representations (for example, GloVe \cite{pennington2014glove})  into sentence representations is an active research topic. Different from previous studies (for example, doc2vec \cite{mikolov2013efficient}, skip-thought vectors \cite{kiros2015skip}, Siamese CBOW \cite{kenter2016siamese}), our main contribution is to represent sentences using {\em non-vector} space representations: a sentence can be well represented by the subspace spanned by the context word vectors -- such a method naturally builds on any  word representation method. Due to the widespread use of word2vec and GloVe, we use their  publicly available word representations -- word2vec\cite{mikolov2013efficient} trained using Google News\footnote{\url{https://code.google.com/archive/p/word2vec/}} and GloVe \cite{pennington2014glove} trained using Common Crawl\footnote{\url{http://nlp.stanford.edu/projects/glove/}} -- to test our observations.

\begin{figure}[!h]
\centering
\includegraphics[width=0.4\textwidth]{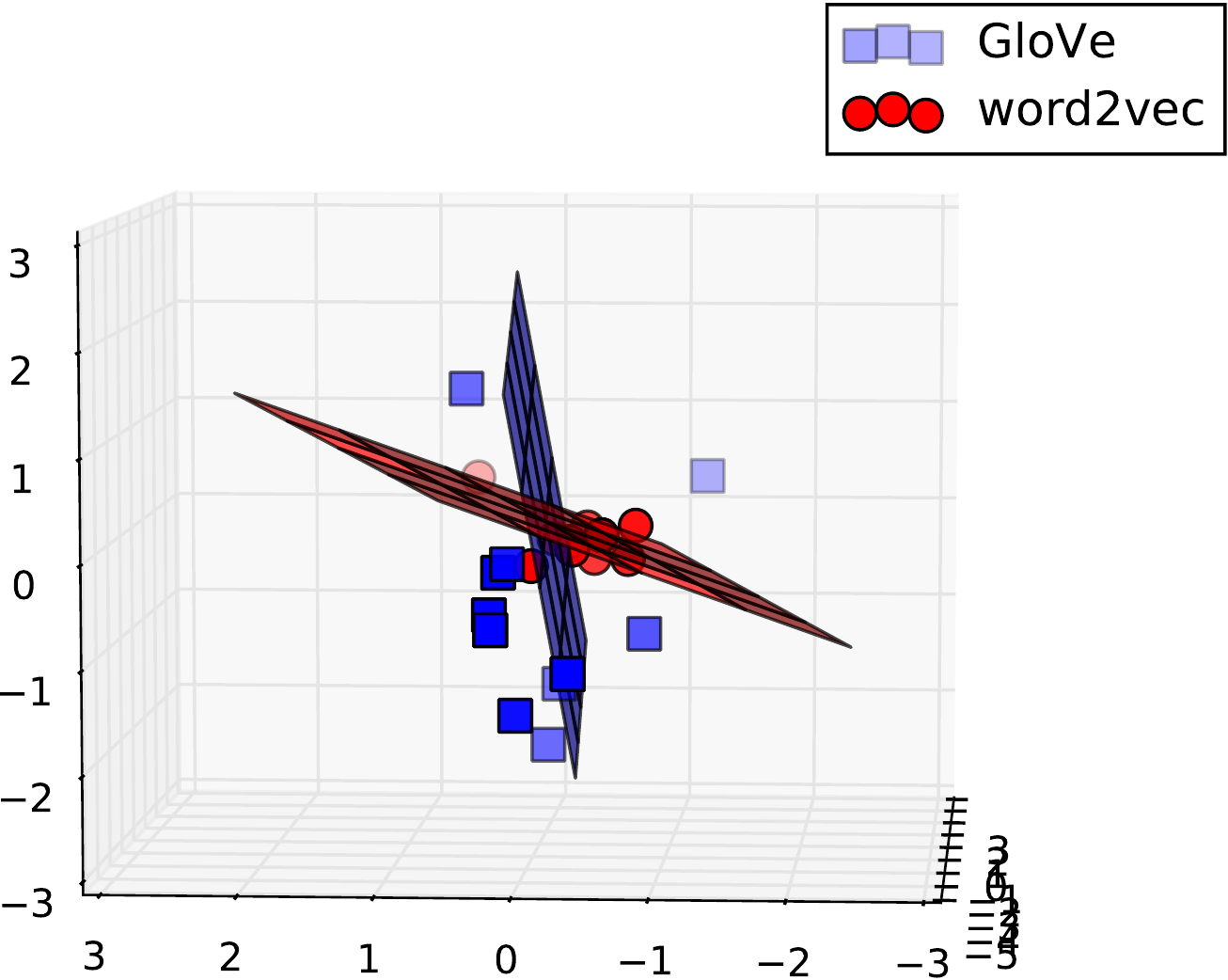}
\caption{Geometry of sentences.}
\label{fig:lowrank-visual}
\end{figure}

\paragraph{Observation} Let $v(w)\in \mathbb{R}^d$ be the $d$-dimensional word representation for a given word $w\in V$, and $s = (w_1,\ldots,w_n)$ be a given sentence. Consider the following sentence where $n=32$: 
\begin{quote}
They would not tell me if there was any pension left here, and would only tell me if there was (and how much there was) if they saw I was entitled to it.
\end{quote}
After stacking the (non-functional) word vectors $v(w)$ to form a $d\times n$ matrix, we observe that most energy (80\% for GloVe and 72\% for word2vec) of such a matrix is contained in a rank-$N$ subspace, where $N$ is much smaller than $n$ (for comparison, we choose $N$ to be 4 and therefore $N/n\approx 13\%$). Figure~\ref{fig:lowrank-visual} provides a visual representation of this geometric phenomenon, where we have projected the $d$-dimensional word representations into 3-dimensional vectors and use these 3-dimensional word vectors to get the subspace for this sentence (we set $N=2$ here for visualization), and plot the subspaces as 2-dimensional planes.

\paragraph{Geometry of Sentences} The example above generalizes to a vast majority of the sentences: the word representations of a given sentence roughly reside in a low-rank subspace, which can be extracted by principal component analysis (PCA). 

\begin{figure}[!h]
\centering
\includegraphics[width=0.4\textwidth]{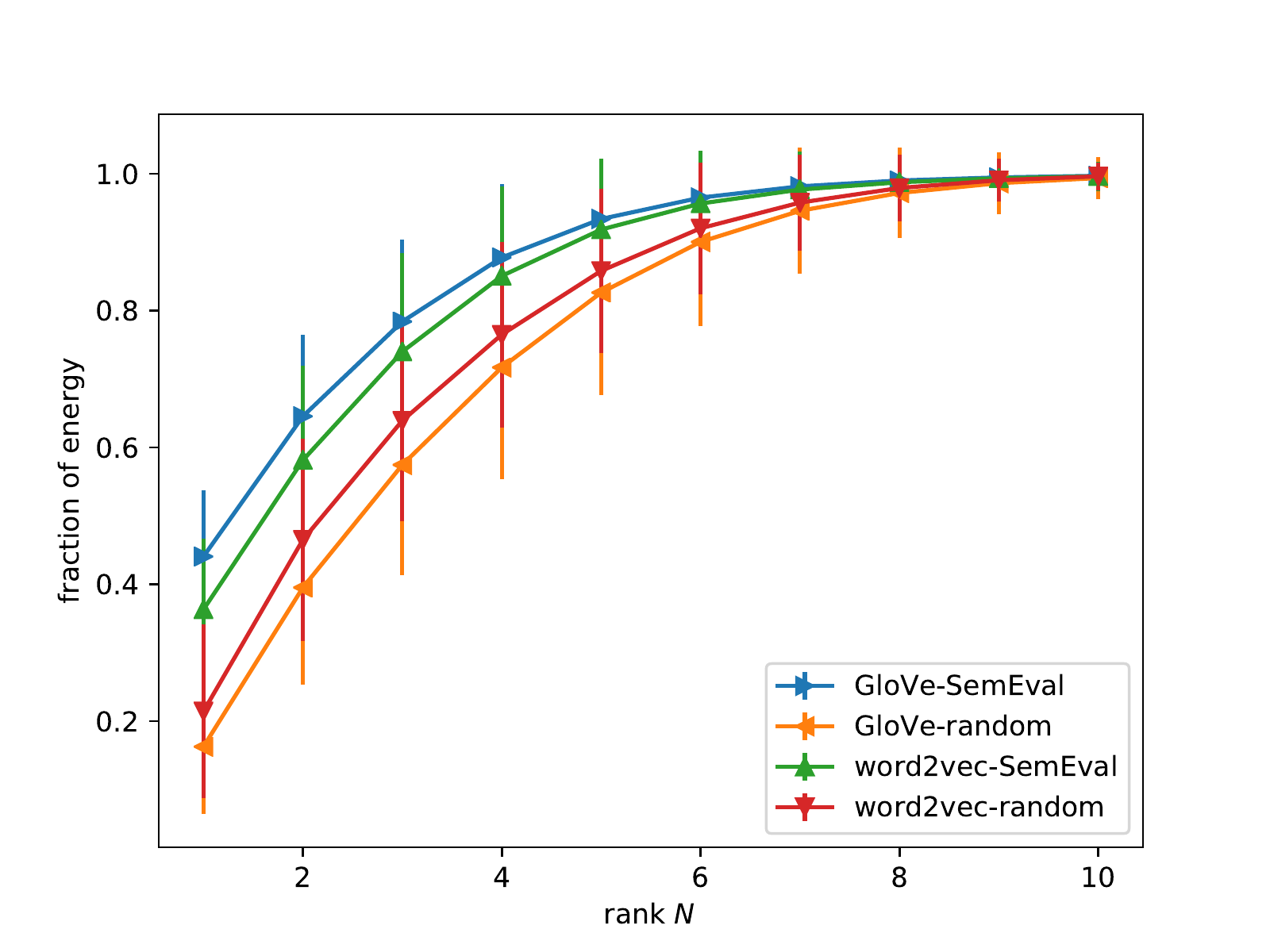}
\caption{Fraction of energy captured by the top principal components.}
\label{fig:variance-ratio}
\end{figure}

\paragraph{Verification} We empirically validate this geometric phenomenon by collecting 53,396 sentences from the SemEval STS share tasks \cite{agirre2012semeval,agirre2013sem,agirre2014semeval,agirrea2015semeval} and plotting the fraction of energy being captured by the top $N$ components of PCA in Figure~\ref{fig:variance-ratio} for $N=3,4,5$, from where we can observe that on average 70\% of the energy is captured by a rank-3 subspace, and 80\% for a rank-4 subspace and 90\% for rank-5 subspace.  For comparison, the fraction of energy of random sentences (generated i.i.d. from the unigram distribution) are also plotted in Figure~\ref{fig:variance-ratio}.

\paragraph{Representation} The observation above motivates our sentence representation algorithm: since the words in a sentence concentrate on a few directions, the subspace spanned by these directions could in principle be a proper representation for this sentence.  The direction and subspace in turn can be extracted via PCA as in Algorithm~\ref{algo:subspace}.

\begin{algorithm}[!h]
\SetKwInOut{Input}{Input}
\SetKwInOut{Output}{Output}
\Input{a sentence $s$, word embeddings $v(\cdot)$, and a PCA rank $N$. }
Compute the first $N$ principle components of samples ${v(w'), w'\in c}$,
\begin{align*}
  u_1,...,u_N &\leftarrow {\rm PCA}({v(w'), w'\in s}), \\
  S &\leftarrow \left\{\sum_{n=1}^N : \alpha_n u_n, \alpha_n \in R\right\}
\end{align*} \\
\Output{$N$ orthonormal basis $u_1, ..., u_N$ and a subspace $S$.}
\caption{The algorithm for sentence representations.}
\label{algo:subspace}
\end{algorithm}

\paragraph{Similarity Metric} The principle of sentence representations is that similar sentences should have similar representations. In this case, we expect the similarity between subspaces to be a good index for the semantic similarity of sentences. In our paper, we define the similarity between subspaces as follows: let $u_1(s),...,u_N(s)$ be the $N$ orthonormal basis for a sentence $s$. After stacking the $N$ vectors in a $d\times N$ matrix $U(s) = (u_1(s),...,u_N(s))$, we define the corresponding cosine similarity as,
$
	{\rm CosSim}(s_1,s_2) = \sqrt{\sum_{t=1}^N \sigma_t^2},
$
where $\sigma_t$ is the $t$-th singular value of $U(s_1)^{\rm T} U(s_2)$.

Note that $\sigma_t = \cos(\theta_t)$ where $\theta_t$ is the $t$-th ``principle angle" between two subspaces. Such a metric is naturally related to the cosine similarity between vectors, which has been empirically validated to be a good measure of word similarity.

\section{Experiments}

\begin{table*}[!h]
\centering
\begin{tabular}{|c|c|c|c|c|c|c|c|c|c|}
\hline
\multicolumn{2}{|c|}{\multirow{2}{*}{dataset}} & \multirow{2}{*}{$l$} & \multirow{2}{*}{ST} & \multirow{2}{*}{SC} & \multirow{2}{*}{D2V} & \multicolumn{2}{c|}{GloVe}                         & \multicolumn{2}{c|}{skip-gram}                         \\ \cline{7-10} 
\multicolumn{2}{|c|}{}    &                      &                     &                     &                          & avg.                            & subspace                        & avg.                 & subspace                        \\ \hline
\multirow{5}{*}{2012}     & MSRpar             & 17.70                & 5.60                & 43.79               & 14.85                    & \underline{\textbf{46.18}}  & 40.74                           & 16.82                & 35.71                           \\ 
                          & MSRvid             & 6.63                 & 58.07               & 45.22               & 19.82                    & \underline{63.75}             & { \textbf{73.90}}           & \underline{58.28}  & \underline{ 63.19}            \\ 
                          & OnWn               & 7.57                 & 60.45               & \textbf{64.44}    & 35.73                    & 56.72                           & 63.21                           & 42.22                & 58.43                           \\ 
                          & SMTeuroparl        & 10.70                & 42.03               & 45.03               & 36.18                    & \underline{ \textbf{52.51}} & \underline{ 45.83}            & 37.99                & \underline{ 45.35}            \\ 
                          & SMTnews            & 11.72                & 39.11               & 39.02               & 52.78                    & 38.99                           & \underline{ \textbf{45.73}} & 23.44                & 37.73                           \\ \hline
\multirow{3}{*}{2013}     & FNWN               & 19.90                & 31.24               & 23.22               & \textbf{51.07}           & 39.29                           & 41.03                           & 19.35                & 26.43                           \\ 
                          & OnWn               & 7.17                 & 24.18               & 49.85               & 49.26                    & \underline{ 52.48}            & \underline{ \textbf{72.03}} & \underline{ 58.30} & \underline{ 56.52}            \\ 
                          & headlines          & 7.21                 & 38.61               & 65.34               & 28.90                    & 49.07                           & \underline{ \textbf{66.13}} & 41.53                & 62.84                           \\ \hline
\multirow{6}{*}{2014}     & OnWn               & 7.74                 & 48.82               & 60.73               & 60.84                    & 60.15                           & \underline{ \textbf{76.28}} & 55.38                & \underline{ 67.13}            \\ 
                          & deft-forum         & 8.38                 & 37.36               & 40.82               & 22.63                    & 22.75                           & \underline{ 42.60}            & 32.87                & \underline{ \textbf{45.30}} \\ 
                          & deft-news          & 15.78                & 46.17               & 59.13               & 18.93                    & \underline{ 62.91}            & \underline{ \textbf{64.40}} & 38.72                & 53.62                           \\ 
                          & headlines          & 7.43                 & 40.31               & \textbf{63.64}    & 24.31                    & 46.00                           & 62.42                           & 36.46                & 61.44                           \\ 
                          & images             & 9.12                 & 42.57               & 64.97               & 39.92                    & 55.19                           & \underline{ \textbf{73.38}} & 45.17                & \underline{ \textbf{71.84}} \\ 
                          & tweet-news         & 10.03                & 51.38               & 73.15               & 33.56                    & 60.45                           & \underline{ \textbf{74.29}} & 44.16                & \underline{ 73.87}            \\ \hline
\multirow{5}{*}{2015}     & answer-forum       & 15.03                & 27.84               & 21.81               & 28.59                    & \underline{ 31.39}            & \underline{ \textbf{69.50}} & \underline{ 34.83} & \underline{ 57.62}            \\ 
                          & answer-studetns    & 10.44                & 26.61               & 36.71               & 11.14                    & \underline{ 48.46}            & \underline{ \textbf{63.43}} & \underline{ 43.85} & \underline{ 59.01}            \\ 
                          & belief             & 13.00                & 45.84               & 47.69               & 30.58                    & 44.73                           & \underline{ \textbf{69.65}} & \underline{ 49.24} & \underline{ 64.48}            \\ 
                          & headlines          & 7.50                 & 12.48               & 21.51               & 22.64                    & \underline{ 44.80}            & \underline{ 65.67}            & \underline{ 44.23} & \underline{ \textbf{68.02}} \\ 
                          & images             & 9.12                 & 21.00               & 25.60               & 34.14                    & \underline{ 66.40}            & \underline{ \textbf{80.12}} & \underline{ 56.47} & \underline{ 70.53}            \\ \hline
\end{tabular}
\caption{Pearson's correlation (x100) on SemEval textual similarity task using 19 different datasets, where $l$ is the average sentence length of each dataset. Results that are better than the baselines are marked with underlines and the best results are in bold.}
\label{tb:sts}
\end{table*}

In this section we evaluate our  sentence representations empirically on the STS tasks. The objective of this task is to test the degree to which the algorithm can capture the semantic equivalence between two sentences. For example, the similarity between ``a kid jumping a ledge with a bike'' and ``a black and white cat playing with a blanket'' is 0 and the similarity between ``a cat standing on tree branches'' and ``a black and white cat is high up on tree branches'' is 3.6. The algorithm is then evaluated in terms of Pearson's correlation between the predicted score and the human judgment.

\paragraph{Datasets} We test the performances of our algorithm on 19 different datasets, which include SemEval STS share tasks \cite{agirre2012semeval,agirre2013sem,agirre2014semeval,agirrea2015semeval}, sourced from multiple domains (for example, News, WordNet definitions and Twitter). 

\paragraph{Baselines and Preliminaries} Our main comparisons are with algorithms that perform unsupervised sentence representation:   average of word representations (i.e., avg. (of GloVe and skip-gram) where we use the average of word vectors), doc2vec (D2V) \cite{le2014distributed} and sophisticated neural networks (i.e., skip-thought vectors (ST) \cite{kiros2015skip}, Siamese CBOW (SC) \cite{kenter2016siamese}). In order to enable a fair comparison, we use the Toronto Book Corpus \cite{moviebook} to train word embeddings. In our experiment, we adapt the same setting as in \cite{kenter2016siamese} where we use skip-gram \cite{mikolov2013efficient} of and GloVe \cite{pennington2014glove} to train a 300-dimensional word vectors for the words that occur 5 times or more in the training corpus. The rank of subspaces is set to be 4 for both word2vec and GloVe.

\paragraph{Results} The detailed results are reported in Table~\ref{tb:sts}, from where we can observe two phenomena: (a) representing sentences by its averaged word vectors provides a strong baseline and the performances are remarkably stable across different datasets; (b) our subspace-based method outperforms the average-based method by 14\% on average and the neural network based approaches by 15\%. This suggests that representing sentences by subspaces maintains more information than simply taking the average, and is more robust than highly-tuned sophisticated models. 

When we average over the words, the average vector is biased because of many irrelevant words (for example, function words) in a given sentence. Therefore, given a longer sentence, the effect of useful word vectors become smaller and thus the average vector is less reliable at capturing the semantics. On the other hand,  the subspace representation is  immune to this phenomenon: the word vectors capturing the semantics of the sentence tend to concentrate on a few directions which dominate the subspace representation.

\section{Conclusion}

This paper presents a novel unsupervised sentence representation leveraging the Grassmannian geometry of word representations. While the current approach relies on the pre-trained word representations, the {\em joint learning}  of both word and sentence representations and in conjunction with supervised datasets such the Paraphrase Database (PPDB) \cite{ganitkevitch2013ppdb} is left to future research.  Also interesting is the exploration of neural network architectures that operate on  subspaces  (as opposed to vectors), allowing for downstream evaluations of our novel representation. 



\bibliography{sentence}

\begin{thebibliography}{}
\expandafter\ifx\csname natexlab\endcsname\relax\def\natexlab#1{#1}\fi

\bibitem[{Adi et~al.(2016)Adi, Kermany, Belinkov, Lavi, and
  Goldberg}]{adi2016fine}
Yossi Adi, Einat Kermany, Yonatan Belinkov, Ofer Lavi, and Yoav Goldberg. 2016.
\newblock Fine-grained analysis of sentence embeddings using auxiliary
  prediction tasks.
\newblock {\em arXiv preprint arXiv:1608.04207\/} .

\bibitem[{Agirre et~al.(2015)Agirre, Banea, Cardie, Cer, Diab, Gonzalez-Agirre,
  Guo, Lopez-Gazpio, Maritxalar, Mihalcea et~al.}]{agirrea2015semeval}
Eneko Agirre, Carmen Banea, Claire Cardie, Daniel Cer, Mona Diab, Aitor
  Gonzalez-Agirre, Weiwei Guo, Inigo Lopez-Gazpio, Montse Maritxalar, Rada
  Mihalcea, et~al. 2015.
\newblock Semeval-2015 task 2: Semantic textual similarity, english, spanish
  and pilot on interpretability.
\newblock In {\em Proceedings of the 9th international workshop on semantic
  evaluation (SemEval 2015)\/}. pages 252--263.

\bibitem[{Agirre et~al.(2014)Agirre, Banea, Cardie, Cer, Diab, Gonzalez-Agirre,
  Guo, Mihalcea, Rigau, and Wiebe}]{agirre2014semeval}
Eneko Agirre, Carmen Banea, Claire Cardie, Daniel Cer, Mona Diab, Aitor
  Gonzalez-Agirre, Weiwei Guo, Rada Mihalcea, German Rigau, and Janyce Wiebe.
  2014.
\newblock Semeval-2014 task 10: Multilingual semantic textual similarity.
\newblock In {\em Proceedings of the 8th international workshop on semantic
  evaluation (SemEval 2014)\/}. pages 81--91.

\bibitem[{Agirre et~al.(2013)Agirre, Cer, Diab, Gonzalez-Agirre, and
  Guo}]{agirre2013sem}
Eneko Agirre, Daniel Cer, Mona Diab, Aitor Gonzalez-Agirre, and Weiwei Guo.
  2013.
\newblock sem 2013 shared task: Semantic textual similarity, including a pilot
  on typed-similarity.
\newblock In {\em In* SEM 2013: The Second Joint Conference on Lexical and
  Computational Semantics. Association for Computational Linguistics\/}.
  Citeseer.

\bibitem[{Agirre et~al.(2012)Agirre, Diab, Cer, and
  Gonzalez-Agirre}]{agirre2012semeval}
Eneko Agirre, Mona Diab, Daniel Cer, and Aitor Gonzalez-Agirre. 2012.
\newblock Semeval-2012 task 6: A pilot on semantic textual similarity.
\newblock In {\em Proceedings of the First Joint Conference on Lexical and
  Computational Semantics-Volume 1: Proceedings of the main conference and the
  shared task, and Volume 2: Proceedings of the Sixth International Workshop on
  Semantic Evaluation\/}. Association for Computational Linguistics, pages
  385--393.

\bibitem[{Arora et~al.(2015)Arora, Li, Liang, Ma, and Risteski}]{arora2015rand}
Sanjeev Arora, Yuanzhi Li, Yingyu Liang, Tengyu Ma, and Andrej Risteski. 2015.
\newblock Rand-walk: A latent variable model approach to word embeddings.
\newblock {\em arXiv preprint arXiv:1502.03520\/} .

\bibitem[{Bengio et~al.(2003)Bengio, Ducharme, Vincent, and
  Jauvin}]{bengio2003neural}
Yoshua Bengio, R{\'e}jean Ducharme, Pascal Vincent, and Christian Jauvin. 2003.
\newblock A neural probabilistic language model.
\newblock {\em journal of machine learning research\/} 3(Feb):1137--1155.

\bibitem[{Collobert et~al.(2011)Collobert, Weston, Bottou, Karlen, Kavukcuoglu,
  and Kuksa}]{collobert2011natural}
Ronan Collobert, Jason Weston, L{\'e}on Bottou, Michael Karlen, Koray
  Kavukcuoglu, and Pavel Kuksa. 2011.
\newblock Natural language processing (almost) from scratch.
\newblock {\em Journal of Machine Learning Research\/} 12(Aug):2493--2537.

\bibitem[{Dhillon et~al.(2012)Dhillon, Rodu, Foster, and
  Ungar}]{dhillon2012two}
Paramveer Dhillon, Jordan Rodu, Dean Foster, and Lyle Ungar. 2012.
\newblock Two step cca: A new spectral method for estimating vector models of
  words.
\newblock {\em arXiv preprint arXiv:1206.6403\/} .

\bibitem[{Ganitkevitch et~al.(2013)Ganitkevitch, {Van Durme}, and
  Callison-Burch}]{ganitkevitch2013ppdb}
Juri Ganitkevitch, Benjamin {Van Durme}, and Chris Callison-Burch. 2013.
\newblock \href{http://cs.jhu.edu/~ccb/publications/ppdb.pdf}{{PPDB}: The
  paraphrase database}.
\newblock In {\em Proceedings of NAACL-HLT\/}. Association for Computational
  Linguistics, Atlanta, Georgia, pages 758--764.
\newblock
  \href{http://cs.jhu.edu/~ccb/publications/ppdb.pdf}{http://cs.jhu.edu/~ccb/publications/ppdb.pdf}.

\bibitem[{Gong et~al.(2017)Gong, Bhat, and Viswanath}]{gong2016geometry}
Hongyu Gong, Suma Bhat, and Pramod Viswanath. 2017.
\newblock Geometry of compositionality.
\newblock {\em Association for Advancement of Artificial Intelligence (AAAI)\/}
  .

\bibitem[{Hill et~al.(2016)Hill, Cho, and Korhonen}]{hill2016learning}
Felix Hill, Kyunghyun Cho, and Anna Korhonen. 2016.
\newblock Learning distributed representations of sentences from unlabelled
  data.
\newblock {\em arXiv preprint arXiv:1602.03483\/} .

\bibitem[{Huang et~al.(2012)Huang, Socher, Manning, and
  Ng}]{huang2012improving}
Eric~H Huang, Richard Socher, Christopher~D Manning, and Andrew~Y Ng. 2012.
\newblock Improving word representations via global context and multiple word
  prototypes.
\newblock In {\em Proceedings of the 50th Annual Meeting of the Association for
  Computational Linguistics: Long Papers-Volume 1\/}. Association for
  Computational Linguistics, pages 873--882.

\bibitem[{Kalchbrenner et~al.(2014)Kalchbrenner, Grefenstette, and
  Blunsom}]{kalchbrenner2014convolutional}
Nal Kalchbrenner, Edward Grefenstette, and Phil Blunsom. 2014.
\newblock A convolutional neural network for modelling sentences.
\newblock {\em arXiv preprint arXiv:1404.2188\/} .

\bibitem[{Kenter et~al.(2016)Kenter, Borisov, and de~Rijke}]{kenter2016siamese}
Tom Kenter, Alexey Borisov, and Maarten de~Rijke. 2016.
\newblock Siamese cbow: Optimizing word embeddings for sentence
  representations.
\newblock {\em arXiv preprint arXiv:1606.04640\/} .

\bibitem[{Kim(2014)}]{kim2014convolutional}
Yoon Kim. 2014.
\newblock Convolutional neural networks for sentence classification.
\newblock {\em arXiv preprint arXiv:1408.5882\/} .

\bibitem[{Kiros et~al.(2015)Kiros, Zhu, Salakhutdinov, Zemel, Urtasun,
  Torralba, and Fidler}]{kiros2015skip}
Ryan Kiros, Yukun Zhu, Ruslan~R Salakhutdinov, Richard Zemel, Raquel Urtasun,
  Antonio Torralba, and Sanja Fidler. 2015.
\newblock Skip-thought vectors.
\newblock In {\em Advances in neural information processing systems\/}. pages
  3294--3302.

\bibitem[{Le and Mikolov(2014)}]{le2014distributed}
Quoc~V Le and Tomas Mikolov. 2014.
\newblock Distributed representations of sentences and documents.
\newblock In {\em ICML\/}. volume~14, pages 1188--1196.

\bibitem[{Levy and Goldberg(2014)}]{levy2014neural}
Omer Levy and Yoav Goldberg. 2014.
\newblock Neural word embedding as implicit matrix factorization.
\newblock In {\em Advances in neural information processing systems\/}. pages
  2177--2185.

\bibitem[{Mikolov et~al.(2013)Mikolov, Chen, Corrado, and
  Dean}]{mikolov2013efficient}
Tomas Mikolov, Kai Chen, Greg Corrado, and Jeffrey Dean. 2013.
\newblock Efficient estimation of word representations in vector space.
\newblock {\em arXiv preprint arXiv:1301.3781\/} .

\bibitem[{Mikolov et~al.(2010)Mikolov, Karafi{\'a}t, Burget, Cernock{\`y}, and
  Khudanpur}]{mikolov2010recurrent}
Tomas Mikolov, Martin Karafi{\'a}t, Lukas Burget, Jan Cernock{\`y}, and Sanjeev
  Khudanpur. 2010.
\newblock Recurrent neural network based language model.
\newblock In {\em Interspeech\/}. volume~2, page~3.

\bibitem[{Mnih and Hinton(2007)}]{mnih2007three}
Andriy Mnih and Geoffrey Hinton. 2007.
\newblock Three new graphical models for statistical language modelling.
\newblock In {\em Proceedings of the 24th international conference on Machine
  learning\/}. ACM, pages 641--648.

\bibitem[{Mu et~al.(2016)Mu, Bhat, and Viswanath}]{mu2016geometry}
Jiaqi Mu, Suma Bhat, and Pramod Viswanath. 2016.
\newblock Geometry of polysemy.
\newblock {\em arXiv preprint arXiv:1610.07569\/} .

\bibitem[{Pennington et~al.(2014)Pennington, Socher, and
  Manning}]{pennington2014glove}
Jeffrey Pennington, Richard Socher, and Christopher~D Manning. 2014.
\newblock Glove: Global vectors for word representation.
\newblock In {\em EMNLP\/}. volume~14, pages 1532--43.

\bibitem[{Socher et~al.(2013)Socher, Perelygin, Wu, Chuang, Manning, Ng, and
  Potts}]{socher2013recursive}
Richard Socher, Alex Perelygin, Jean~Y Wu, Jason Chuang, Christopher~D Manning,
  Andrew~Y Ng, and Christopher Potts. 2013.
\newblock Recursive deep models for semantic compositionality over a sentiment
  treebank.
\newblock In {\em Proceedings of the conference on empirical methods in natural
  language processing (EMNLP)\/}. Citeseer, volume 1631, page 1642.

\bibitem[{Stratos et~al.(2015)Stratos, Collins, and Hsu}]{stratos2015model}
Karl Stratos, Michael Collins, and Daniel Hsu. 2015.
\newblock Model-based word embeddings from decompositions of count matrices.
\newblock In {\em Proceedings of ACL\/}. pages 1282--1291.

\bibitem[{Sutskever et~al.(2014)Sutskever, Vinyals, and
  Le}]{sutskever2014sequence}
Ilya Sutskever, Oriol Vinyals, and Quoc~V Le. 2014.
\newblock Sequence to sequence learning with neural networks.
\newblock In {\em Advances in neural information processing systems\/}. pages
  3104--3112.

\bibitem[{Wieting et~al.(2015)Wieting, Bansal, Gimpel, and
  Livescu}]{wieting2015towards}
John Wieting, Mohit Bansal, Kevin Gimpel, and Karen Livescu. 2015.
\newblock Towards universal paraphrastic sentence embeddings.
\newblock {\em arXiv preprint arXiv:1511.08198\/} .

\bibitem[{Zhu et~al.(2015)Zhu, Kiros, Zemel, Salakhutdinov, Urtasun, Torralba,
  and Fidler}]{moviebook}
Yukun Zhu, Ryan Kiros, Richard Zemel, Ruslan Salakhutdinov, Raquel Urtasun,
  Antonio Torralba, and Sanja Fidler. 2015.
\newblock Aligning books and movies: Towards story-like visual explanations by
  watching movies and reading books.
\newblock In {\em arXiv preprint arXiv:1506.06724\/}.

\end{thebibliography}
\bibliographystyle{acl_natbib}

\end{document}